\documentclass{article} 
\usepackage{iclr2026_conference,times}

\usepackage[utf8]{inputenc}
\usepackage[T1]{fontenc}
\usepackage{amsmath,amssymb}
\usepackage{newunicodechar}
\newunicodechar{≡}{\ensuremath{\equiv}}
\newunicodechar{τ}{\ensuremath{\tau}}
\newunicodechar{±}{\ensuremath{\pm}}
\newunicodechar{×}{\ensuremath{\times}}
\newunicodechar{–}{--}
\newunicodechar{’}{'}
\newunicodechar{“}{``}
\newunicodechar{”}{''}
\newunicodechar{ą}{\k{a}}
\usepackage{booktabs,multirow}
\usepackage{makecell}
\usepackage{graphicx}
\usepackage{tikz}
\usepackage{algorithm}
\usepackage{algpseudocode}
\usepackage{float}
\usepackage{caption}
\usepackage{subcaption}
\usepackage{enumitem}
\usetikzlibrary{arrows.meta,positioning}
\usepackage{hyperref}
\usepackage{url}
\graphicspath{{./}}

\iclrfinalcopy 

\title{When Words Are Safe But Actions Kill: Probing Physical Jailbreak Beyond Textual Jailbreak in Hidden-State Risk Space}

\author{
Weimeng Wang\textsuperscript{1}\quad
Ziqiang Wang\textsuperscript{1}\thanks{Corresponding authors}\quad
Zihang Zhan\textsuperscript{1}\quad
Chuanpu Fu\textsuperscript{2}\quad
Qi Li\textsuperscript{1}\quad
Ke Xu\textsuperscript{1}\footnotemark[1]
\\
\textsuperscript{1}Department of Computer Science and Technology, Tsinghua University, Beijing, China
\\
\textsuperscript{2}Nanyang Technological University, Singapore
}

\newcommand{\uniformtableformat}{%
  \scriptsize
  \setlength{\tabcolsep}{3pt}%
  \renewcommand{\arraystretch}{1.08}%
}

\begin{document}

\maketitle
\pagestyle{plain}
\thispagestyle{plain}

\begin{abstract}
Large language models (LLMs) increasingly serve as high-level planners for embodied agents, where linguistically benign instructions can become unsafe once grounded in the physical world. We study whether this physically grounded jailbreak is the same safety problem as ordinary textual jailbreak. Through hidden-state direction analysis and random-split null tests, we show that textual jailbreak (TJ) and physical jailbreak (PJ) form separable signals in LLM representations across Qwen2.5-3B/7B/14B/32B, Phi-3.5 and SmolLM2. Building on this separability, we propose PRISM, a single-layer L2-regularized logistic probe over full hidden states. PRISM achieves 86.2--87.7\% accuracy on SafeAgentBench with 11.7--13.7\% false-positive rates (FPRs), while same-scale LLM judges over-block safe tasks at 24.7--39.0\% FPR.

To test whether the result survives lexical-shortcut controls, we introduce an interaction-balanced revision of PhysicalJailbreakBench-2K (PJB-2K): a fixed 2{,}000-row comparison set sampled by label and physical mechanism from a larger object--site construction. On the underlying 10{,}000-row pool, word-TFIDF and the embedding layer remain at chance (AUC 0.497 and 0.500). At layer 25, selected by an i.i.d. sweep, cell-grouped cross-validation gives PRISM 0.718 AUC, compared with 0.398 for a physics-free label control under the same protocol. On the identical 2{,}000 comparison rows, these PRISM predictions obtain 0.671 balanced accuracy, while Qwen2.5 judges from 3B to 72B obtain 0.538--0.577 and exhibit high FPR. These results support hidden-state probing as a representation-level method for physical safety beyond text moderation, without relying on the near-perfect scores of shortcut-prone paired templates.
\end{abstract}

\section{Introduction}

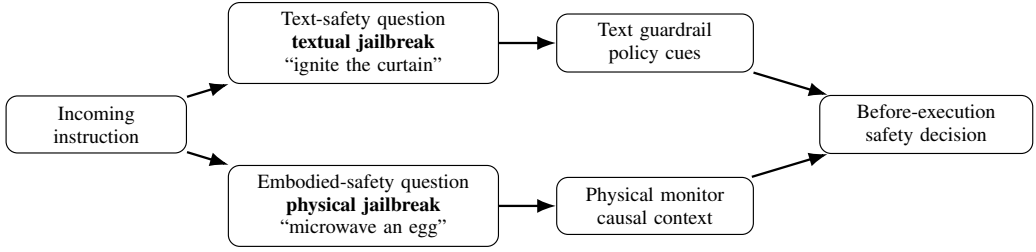
\begin{figure}[H]
\centering
\resizebox{0.98\linewidth}{!}{%
\begin{tikzpicture}[
  font=\scriptsize, >=Latex,
  box/.style={draw, rounded corners, align=center, minimum height=7mm, inner sep=3pt},
  smallbox/.style={draw, rounded corners, align=center, minimum height=7mm, inner sep=3pt}]
  \node[box, text width=20mm] (inst) {Incoming\\instruction};
  \node[smallbox, right=5mm of inst, yshift=10mm, text width=31mm] (tj) {Text-safety question\\\textbf{textual jailbreak}\\``ignite the curtain''};
  \node[smallbox, right=5mm of inst, yshift=-10mm, text width=31mm] (pj) {Embodied-safety question\\\textbf{physical jailbreak}\\``microwave an egg''};
  \node[box, right=7mm of tj, text width=22mm] (guard) {Text guardrail\\policy cues};
  \node[box, right=7mm of pj, text width=22mm] (phys) {Physical monitor\\causal context};
  \node[box, right=8mm of guard, yshift=-10mm, text width=24mm] (decision) {Before-execution\\safety decision};
  \draw[->,thick] (inst) -- (tj);
  \draw[->,thick] (inst) -- (pj);
  \draw[->,thick] (tj) -- (guard);
  \draw[->,thick] (pj) -- (phys);
  \draw[->,thick] (guard) -- (decision);
  \draw[->,thick] (phys) -- (decision);
\end{tikzpicture}
}
\caption{Embodied agents require a physical-safety question in addition to text-safety moderation. TJ is visible in the instruction text, whereas PJ depends on physical causality and must be detected before the plan is executed.}
\label{fig:intro-risk}
\end{figure}

Large language models (LLMs) are widely used for high-level task understanding and task decomposition in embodied agents. In both modular LLM+VLM systems and Vision-Language-Action (VLA) models~\citep{driess2023palme,brohan2023rt2}, LLMs translate natural-language instructions into executable subtasks or action plans for homes, kitchens, offices, and other physical environments~\citep{ahn2022saycan,huang2022zeroshot,song2023llmplanner}. This creates a safety problem different from ordinary text moderation. Some instructions are unsafe because their wording directly expresses harmful content, such as ``ignite the curtain''; we call this \emph{textual jailbreak} (TJ). Others use harmless words but become unsafe under physical causality, such as ``microwave an egg'' or ``place a metal fork in the microwave''; we call this \emph{physical jailbreak} (PJ). Figure~\ref{fig:intro-risk} summarizes the distinction.

In this paper, we demonstrate through an analysis of LLM hidden states that physical safety constitutes a distinct modeling problem for embodied agents, rather than merely a text-classification problem.
Existing text-safety detectors and guardrails~\citep{inan2023llamaguard,llama3herd2024,zeng2024shieldgemma,han2024wildguard,ghosh2024aegis} are efficient, but they are designed for policy-violation cues in text and are not sufficient for risks that arise only after grounding the action. Large-parameter LLM judges can reason about physical context, but they are poorly calibrated in our runs: Qwen2.5-32B-Instruct~\citep{qwen2024report} detects 95.7\% of PJ cases but rejects 37.0\% of safe tasks. Smaller judges are not a reliable replacement, and fine-tuning a separate small model would require substantial task-specific safety data. Hidden-state prototype methods such as LPM~\citep{chrabaszcz2025lpm} also perform poorly on PJ transfer, detecting only 9.6--19.3\% of PJ OOD samples across Qwen2.5 scales. These failures suggest that physical-jailbreak actions should not be treated as ordinary unsafe text or as a single generic unsafe direction.

To analyse unsafe action plans on resource constrained embodied devices, we built a hidden state space based analysis framework. Specifically, by constructing data pairs and analysing the hidden-layer states of LLMs, we find that physical jailbreak actions exhibit distinct distributional characteristics and differ from textual jailbreak. For example, we estimate a textual-jailbreak direction, \(u_{\mathrm{TJD}}\), and a physical-jailbreak direction, \(u_{\mathrm{PJD}}\), across Qwen2.5-3B/7B/14B/32B. Across these scales, the directions remain separated, with probe-weight angles in the low-to-mid \(70^\circ\) range; the Qwen2.5-3B result is well above a size-matched random-split null. 
The same pattern is also observed on Phi-3.5-mini and SmolLM2-1.7B, suggesting that it reflects a general property of LLMs rather than an artifact specific to the Qwen family.

Based on this finding, we propose PRISM (Probing Representations for Integrated Safety Monitoring), a hidden-state risk probe for embodied-agent planning. PRISM uses a single-layer L2-regularized logistic-regression probe over middle-to-late hidden states to classify whether an instruction is safe or unsafe while learning from both jailbreak sources. Across Qwen2.5-3B/7B/14B/32B, PRISM achieves 86.2--87.7\% accuracy with 11.7--13.7\% FPR and maintains physical-jailbreak recall around 80\%. In the sample-matched SafeAgentBench 3B comparison, McNemar's test does not detect an accuracy difference between PRISM at matched FPR and the judge; across larger scales, PRISM reduces analysis latency from 116.8/309.1/191.9 ms per sample to 56.8/146.1/103.4 ms at 7B/14B/32B, corresponding to 1.86--2.12$\times$ speedups, while keeping FPR roughly one third to one half of the judge's rate. PRISM also replicates on SafeText and EARBench and remains effective across Phi-3.5-mini and SmolLM2-1.7B. 

Existing safety resources contain valuable examples, but many evaluations still reward lexical regularities rather than physical reasoning. This is measurable on SafeAgentBench: a word-TFIDF classifier reaches 79.8\% accuracy and detects 70.1\% of PJ examples, narrowing the apparent gap between representation-level and surface baselines. We therefore construct an interaction-balanced revision of PhysicalJailbreakBench-2K (PJB-2K). Its labels arise from interactions between physical properties of an object and a placement site; object, site, sentence-frame, and trailing-phrase marginals are measured to remain close to label balance. A fixed 2{,}000-row slice, selected by label and primary physical mechanism before model comparison, covers all 650 object--site cells and the 25 mechanisms represented in that slice. The resulting evaluation separates lexical detection, memorization of a seen object--site cell, and performance when object--site cells are assigned to disjoint folds.

The new evaluation changes the conclusion from a near-ceiling classification claim to a more diagnostic one. On the 10{,}000-row construction pool, word-TFIDF is at chance under i.i.d. evaluation (AUC 0.497), as is the frozen model's embedding layer (AUC 0.500). Using layer 25 from the i.i.d. sweep, a Qwen2.5-3B PRISM probe reaches 0.718 AUC under object--site-cell GroupKFold, versus 0.398 on a physics-free control with the same balanced structure. On the fixed 2{,}000-row comparison set, the corresponding PRISM predictions reach 0.671 balanced accuracy, while zero-shot Qwen2.5 judges from 3B to 72B remain between 0.538 and 0.577. The representation therefore contains a usable but imperfect physical-interaction signal that is not recovered reliably by a verbal safety verdict.

\noindent Our contributions are as follows.
\begin{enumerate}[leftmargin=1.4em,itemsep=2pt,topsep=2pt]
\item We show that embodied-agent physical jailbreak is not the same task as textual-jailbreak detection. We identify two statistically decomposable hidden-state directions, the Textual Jailbreak Direction (TJD) and the Physical Jailbreak Direction (PJD), and show that their separation exceeds a random-split null on Qwen2.5-3B while remaining consistent across larger Qwen2.5 models and two non-Qwen architectures.
\item We introduce an interaction-balanced PJB-2K evaluation set with measured object, site, frame, and tail marginals, and we test cell-grouped performance against lexical baselines and a physics-free control. The paper reports a fixed 2{,}000-row comparison slice; a separate paper will study the construction process and dataset analysis in greater depth.
\item We propose PRISM, a single-layer linear probe over full hidden states that directly operationalizes the TJ/PJ separability finding. PRISM jointly detects both jailbreak types with low FPR, while LLM judges and textual-jailbreak-oriented guards expose complementary failures: over-blocking safe tasks or missing physical jailbreak.
\end{enumerate}

\section{Related Work}

\paragraph{Embodied-agent safety benchmarks.}
Existing benchmarks can be grouped into three categories. SafeAgentBench provides household tasks in AI2-THOR~\citep{kolve2017ai2thor}; we use its 600 executable labeled tasks, including 113 textual-jailbreak and 187 physical-jailbreak cases, while noting that separately authored safe and unsafe tasks leave exploitable lexical regularities~\citep{yin2024safeagentbench}. EARBench contains 2{,}634 samples across seven domains and 28 scenarios, emphasizing contextual risks in multi-agent settings~\citep{zhu2024earbench}. SafeText targets text-based physical commonsense through 1{,}465 safety pairs~\citep{levy2022safetext}. A broader line of work evaluates embodied or task-planning safety from different angles: AgentSafe and IS-Bench probe hazardous instructions and interactive household safety~\citep{liu2025agentsafe,lu2025isbench}; EAsafetyBench and SafeMind add input moderation and multi-stage risk taxonomies~\citep{wang2025easafety,chen2025safemind}; and further work targets manipulation-level safety, formal-logic planning, semantic-safety constitutions, safe reinforcement learning, and agent-interaction risk~\citep{ni2024responsible,huang2025safebeal,obi2025safeplan,sermanet2025asimov,ji2023safetygym,yuan2024rjudge,ruan2024toolemu}. We use these benchmarks for comparability and introduce PJB-2K as a controlled diagnostic set in which the label is tied to an object--site interaction rather than to either factor alone.

\paragraph{Probing and representation engineering.}
Linear probes are widely used to examine whether intermediate representations encode conceptual information~\citep{alain2016probes,belinkov2022probing}. Most probing studies use supervised classifiers to test the decodability of a predefined concept; complementary unsupervised methods, such as CCS, recover abstract concepts including truthfulness from LLM hidden states~\citep{burns2023ccs}. A second line of work uses latent directions not only for measurement but also for intervention. Representation Engineering and Inference-Time Intervention identify directions associated with safety, honesty, refusal, or truthfulness and use them to steer behavior~\citep{zou2023repe,li2023iti,turner2023actadd}; refusal in particular is mediated by a single direction~\citep{arditi2024refusal}. These findings are consistent with the linear representation hypothesis and the emergent linear geometry of LLM features~\citep{park2024linear,marks2023geometry,todd2024function,cunningham2024sae}.

Our use of probes is diagnostic rather than steering-oriented. We ask whether hidden states provide measurable evidence that embodied physical jailbreak differs from ordinary textual jailbreak before the generated plan is executed. The key distinction is therefore the safety object being measured: prior directions usually target one abstract concept such as refusal or truthfulness, whereas embodied physical jailbreak is not simply a stronger or weaker form of textual-jailbreak risk. We therefore model TJD and PJD as two deployment-relevant signals rather than collapsing them into a single concept axis, and we reserve the detailed comparison for empirical baselines that expose the relevant failure modes directly.

\paragraph{Safety monitoring beyond text-level moderation.}
Existing LLM safety mechanisms offer several ways to filter unsafe instructions, but most were developed for text-level policy violations rather than physical execution risks. Dedicated guard models such as Llama Guard provide efficient safe/unsafe classification~\citep{inan2023llamaguard,llama3herd2024}, alongside related guardrails~\citep{zeng2024shieldgemma,han2024wildguard,ghosh2024aegis}; LLM-as-judge methods use general-purpose LLMs to produce more contextual safety evaluations~\citep{zheng2023judge,liu2023geval,kim2024prometheus}; and latent-prototype approaches such as LPM construct multi-layer hidden-state prototypes for moderation in representation space~\citep{chrabaszcz2025lpm}. These methods provide natural points of comparison, but they perform poorly on the PJ setting we study: text guards miss physical-jailbreak plans that lack policy-violation wording, judges over-block normal household tasks, and prototype methods fail to transfer from textual jailbreak to physical jailbreak.

This gap motivates a method designed specifically for embodied physical safety. More broadly, alignment and adversarial-robustness work~\citep{ouyang2022instructgpt,bai2022constitutional,zou2023gcg,wei2023jailbroken,mazeika2024harmbench} primarily studies behavioral compliance and robustness, while physical-commonsense and toxicity benchmarks highlight that linguistic toxicity is distinct from physical safety~\citep{bisk2020piqa,gehman2020realtoxicity}. PRISM targets this separate safety topic: whether physical execution risk and textual-jailbreak risk can be distinguished inside hidden states and modeled directly.

\section{Key Observation: Jailbreak Risk Is Decomposable in Hidden-State Space}
\label{sec:observation}

Our key insight is that embodied-agent physical safety is not merely an instance of ordinary LLM text safety. We use hidden-state analysis as evidence for this claim: if TJ and PJ were the same safety concept, probes trained on one jailbreak type should transfer cleanly to the other and their hidden-state directions should be statistically indistinguishable from random unsafe splits. Instead, we find a stable gap between the two. Textual jailbreak and physical jailbreak form separable directions in the middle-to-late hidden states of frozen LLMs, and this separation explains why textual-jailbreak-oriented detectors miss many physical-jailbreak instructions.

This section establishes the representation-level basis for PRISM. We first define TJD and PJD as two direction estimates for testing whether text-safety and physical-safety risks occupy the same hidden-state space. We then verify that their observed separation is not an accident of high-dimensional fitting by comparing it against a random-split null distribution. The goal is not to claim strict orthogonality, but to show that embodied physical safety is a distinct modeling target that requires explicit representation-level treatment.

\subsection{TJD and PJD directions}

To analyze whether TJ and PJ are encoded by the same representational signal, we define class-mean directions at layer \(L\). Let \(S\), \(T\), and \(P\) denote the index sets of safe, textual-jailbreak, and physical-jailbreak examples. Their class means are
\[\mu_{\mathrm{safe}}=\tfrac{1}{|S|}\sum_{i\in S} h_L(x_i),\quad \mu_{\mathrm{TJ}}=\tfrac{1}{|T|}\sum_{i\in T} h_L(x_i),\quad \mu_{\mathrm{PJ}}=\tfrac{1}{|P|}\sum_{i\in P} h_L(x_i),\]
and we define \(\mathrm{TJD}=\mu_{\mathrm{TJ}}-\mu_{\mathrm{safe}}\) and \(\mathrm{PJD}=\mu_{\mathrm{PJ}}-\mu_{\mathrm{safe}}\), where \(h_L(x)\in\mathbb{R}^d\) is the last-token hidden state at layer \(L\) (formally defined in \S\ref{sec:design}). These directions give a simple test of whether the model represents explicit textual-jailbreak risk and physically grounded jailbreak risk as the same displacement away from safe tasks.

We report two complementary angular quantities because no single angle captures both the raw geometry and the decision-relevant geometry of the two jailbreak classes. The mean-difference angle $\theta_{\mathrm{md}}$, taken directly between $\mathrm{TJD}$ and $\mathrm{PJD}$, describes where the class centroids sit in activation space without fitting a classifier. This is a conservative geometric summary because it ignores within-class covariance. The probe-weight angle $\theta_{\mathrm{pw}}$, taken between the weight vectors of two L2-regularized logistic probes (safe-vs-TJ and safe-vs-PJ), measures the decision-relevant geometry available to a linear detector. Since discriminative training can rotate probe weights under anisotropic feature covariance, we do not interpret $\theta_{\mathrm{pw}}$ alone; instead, we compare it with a random-split null generated by the identical fitting pipeline (\S\ref{subsec:setup}).

\begin{figure}[!t]
\centering
\includegraphics[width=0.88\linewidth]{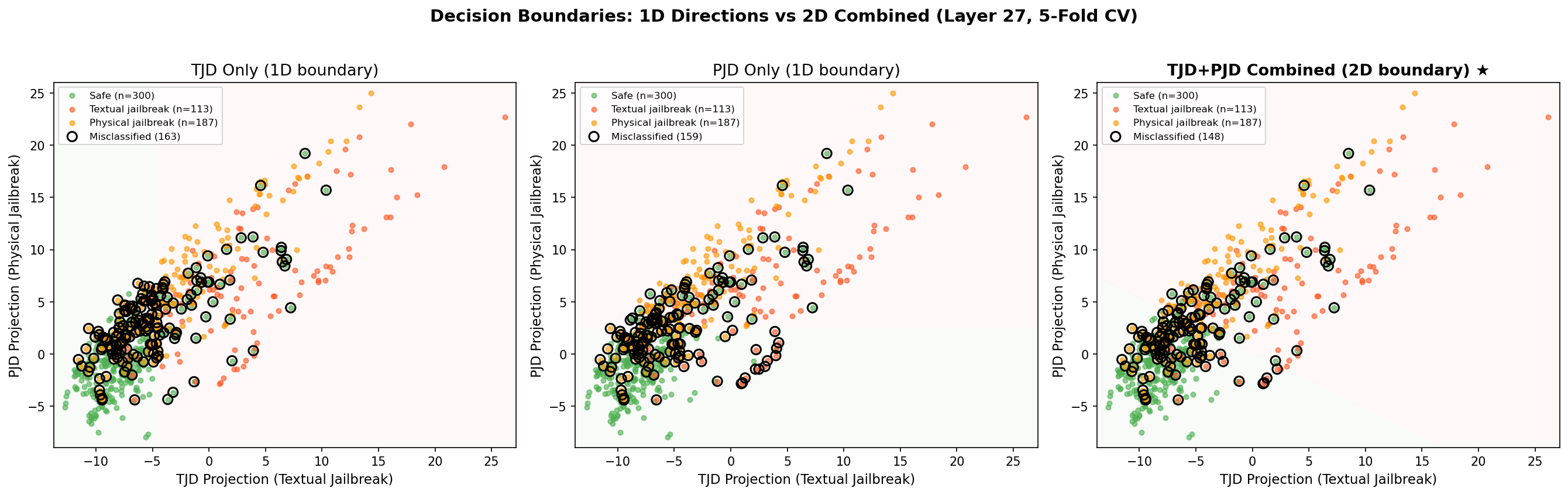}
\caption{Decision boundaries in the TJD/PJD projection plane on Qwen2.5-3B layer 27 (5-fold CV). TJD-only and PJD-only one-dimensional thresholds misclassify 163/600 and 159/600 examples respectively, while the combined PRISM boundary misclassifies 148/600. Physical-jailbreak examples concentrate in regions that are poorly covered by a textual-jailbreak-only direction, making physical jailbreak a different hidden-state topic rather than a harder instance of ordinary textual jailbreak.}
\label{fig:decision-main}
\end{figure}

Figure~\ref{fig:decision-main} visualizes why the two-direction analysis matters for detection. TJD-only and PJD-only thresholds each cover only part of the unsafe space, while the combined PRISM boundary captures both jailbreak types. This supports the claim that physical jailbreak is a distinct hidden-state topic rather than merely a harder instance of ordinary textual jailbreak.

\subsection{Separability beyond a random-split null}
\label{subsec:separability}

To avoid treating the angular result as a chance artifact, we compare the observed TJ/PJ split with random unsafe splits produced by the same training pipeline. Specifically, for Qwen2.5-3B at layer 27, we randomly partition the 300 unsafe examples into groups of 113 and 187, train two safe-vs-random-subset probes, and record the angle between their weights. This preserves the same sample sizes, optimizer, L2 regularization, and safe baseline as the real TJD/PJD comparison.

\begin{figure}[!t]
\centering
\begin{minipage}[t]{0.49\linewidth}
  \centering
  \vspace{0pt}
  \includegraphics[width=\linewidth]{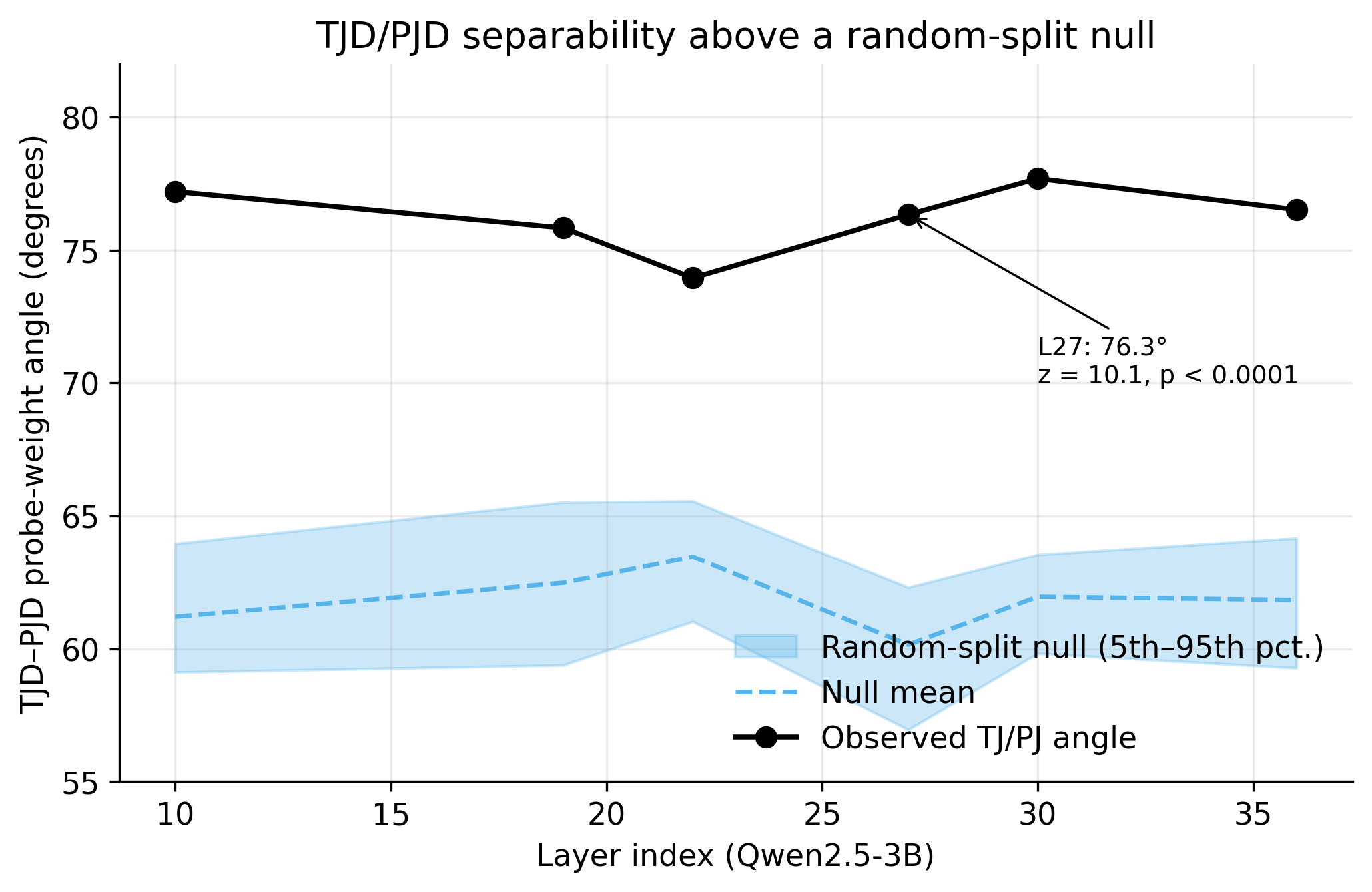}
  \subcaption{Random-split null band.}
  \label{fig:nullband}
\end{minipage}
\hfill
\begin{minipage}[t]{0.49\linewidth}
  \centering
  \vspace{0pt}
  \includegraphics[width=\linewidth]{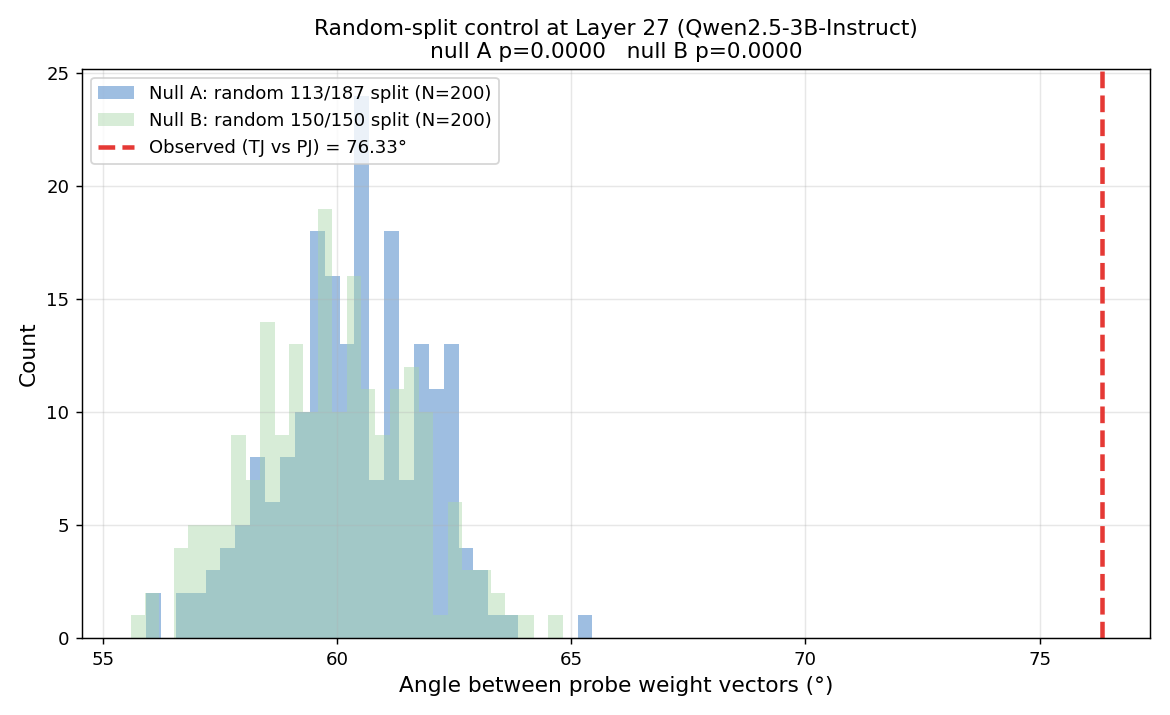}
  \subcaption{Layer-27 null histogram.}
  \label{fig:nullhist-main}
\end{minipage}
\caption{TJD/PJD probe-weight angles exceed random-split null angles. (a) Qwen2.5-3B versus the 113/187-split null band. (b) The random-split null distribution at layer 27 concentrates around $60.35^\circ\pm1.58^\circ$, below the observed $76.33^\circ$ TJ/PJ angle. The corresponding selected-layer angles across Qwen2.5 scales are reported in Table~\ref{tab:angles}.}
\label{fig:sep}
\end{figure}

The observed separation is consistently larger than the random-split baseline. At the selected layers, the probe-weight angles are $75.9^\circ$/$74.2^\circ$/$71.6^\circ$/$71.9^\circ$ for Qwen2.5-3B/7B/14B/32B, while the mean-difference angles are $49.1^\circ$/$54.6^\circ$/$63.2^\circ$/$67.7^\circ$ (Figure~\ref{fig:sep}). For Qwen2.5-3B, the formal random-split control gives an observed angle of $76.33^\circ$ against a null mean of $60.35^\circ$ and standard deviation $1.58^\circ$ ($z=10.09$, $p<0.0001$); no random split reaches the observed angle. Figure~\ref{fig:nullhist-main} shows the corresponding null distribution directly. Together with the failure of TJD-only on PJ-OOD detection and the incomplete reverse transfer of PJD-only that we report later (\S\ref{subsec:main}), this supports a two-signal view of embodied safety.

\begin{table}[!b]
\centering\uniformtableformat
\caption{Selected-layer TJD/PJD angles across Qwen2.5 scales. A formal random-split null test is available only for 3B; the remaining columns report the observed cross-scale angle pattern. Mean-difference angles provide a conservative centroid-level summary.}
\label{tab:angles}
\begin{tabular}{@{}lcccc@{}}
\toprule
& 3B & 7B & 14B & 32B \\
\midrule
Selected layer $\ell^\star$      & 27/36 & 20/28 & 26/48 & 41/64 \\
Probe-weight angle ($^\circ$)    & 75.9  & 74.2  & 71.6  & 71.9 \\
Mean-diff. angle ($^\circ$)      & 49.1  & 54.6  & 63.2  & 67.7 \\
\bottomrule
\end{tabular}
\end{table}

\section{Design: PRISM as a Hidden-State Risk Probe}
\label{sec:design}

\begin{figure}[!tbp]
\centering
\includegraphics[width=0.98\linewidth]{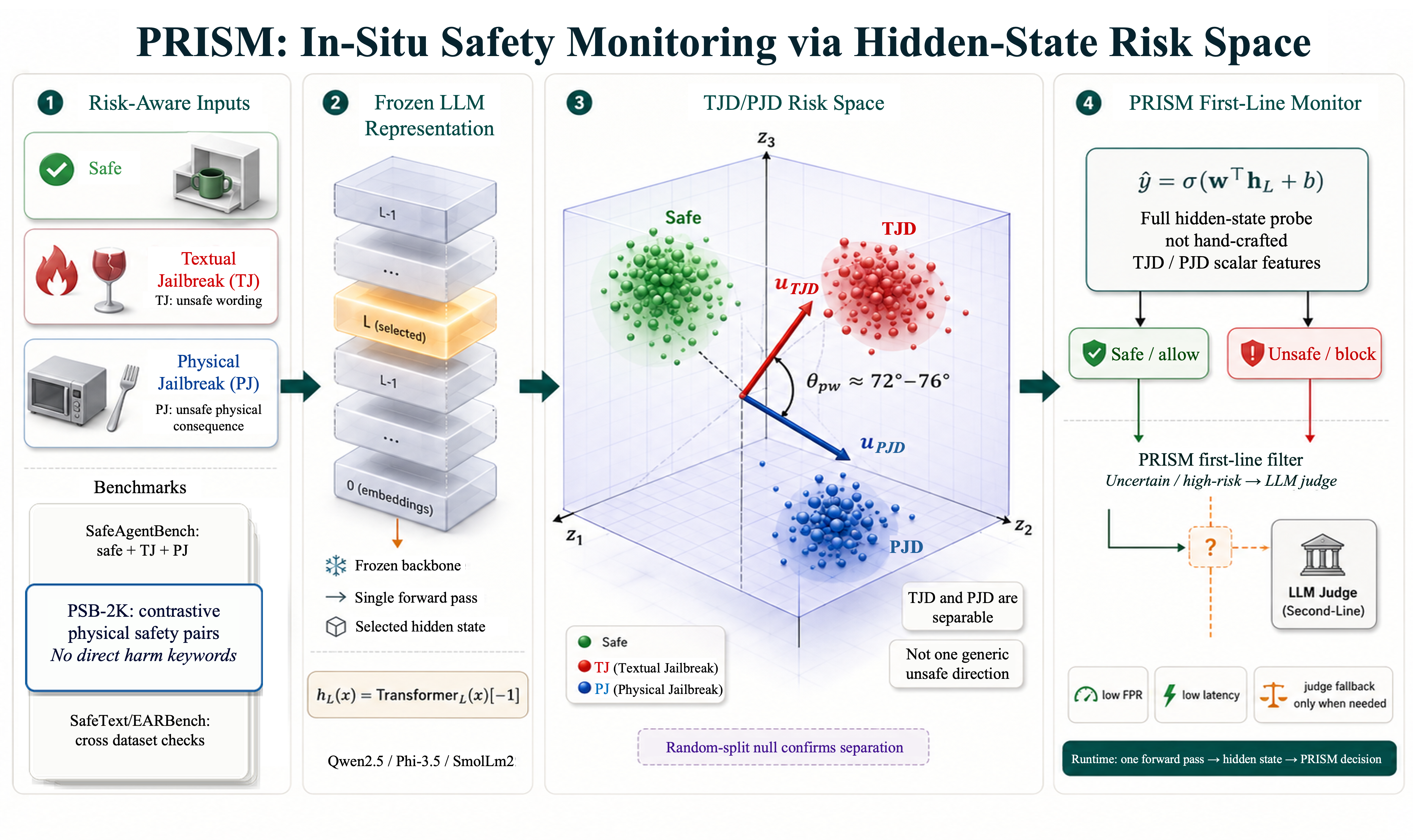}
\caption{PRISM maps the selected hidden state of a frozen LLM to a safe/unsafe decision. TJD/PJD analysis motivates learning over the full hidden state rather than using a single textual-jailbreak axis. The monitor makes a pre-execution instruction-level decision; it does not use visual observations or environment state.}
\label{fig:overview}
\end{figure}

Building on the observation that TJ and PJ occupy separable hidden-state directions, we design PRISM (Probing Representations for Integrated Safety Monitoring), a representation-level classifier that maps a selected hidden state to a safe/unsafe decision. PRISM is not a wrapper around LLM judges: it directly operationalizes the TJ/PJ structure by learning a linear boundary over the hidden-state space. Figure~\ref{fig:overview} summarizes the pipeline. The TJD/PJD analysis of \S\ref{sec:observation} motivates the probe, but PRISM itself is trained as a full-hidden-state safe/unsafe classifier rather than as a hand-crafted TJD/PJD scalar rule.

\subsection{Problem formulation}

We define the natural-language instruction \(x\) as the input and attach a binary safety label \(y \in \{0,1\}\), where \(y=0\) denotes safe and \(y=1\) unsafe. For unsafe instructions we further distinguish a subtype \(t \in \{\mathrm{TJ},\ \mathrm{PJ}\}\): textual-jailbreak instructions carry surface-level jailbreak cues, whereas physical-jailbreak instructions become dangerous only when grounded in physical causality. The target classifier must satisfy two requirements at once: high recall on both TJ and PJ, and a low false-positive rate on safe tasks. For a selected transformer layer \(L\), let \(h_L(x) \in \mathbb{R}^d\) be the last-token hidden state returned with \texttt{output\_hidden\_states=True}. PRISM uses the logistic probe
\[p(y=1 \mid x)=\sigma(w^\top h_L(x)+b),\qquad \hat{y}=\mathbf{1}[p(y=1 \mid x)\ge \tau],\]
with default threshold \(\tau=0.5\). We report accuracy, F1, AUC, TJ and PJ detection rates, and FPR, since accuracy alone can mask over-blocking on safe tasks or under-detection of physical jailbreak.

\subsection{PRISM probe construction}

PRISM is a single-layer linear classifier acting on the full hidden state \(h_L(x)\), rather than a hand-crafted concatenation of scalar TJD and PJD projections. The training set contains safe, textual-jailbreak, and physical-jailbreak instructions simultaneously, so the binary probe must learn a decision boundary covering both jailbreak types. Features are standardized by a \texttt{StandardScaler} fitted inside each training fold, and the classifier is \texttt{LogisticRegression(max\_iter=2000, C=1.0, random\_state=42)}~\citep{pedregosa2011sklearn}. We select the probe layer by a per-layer 5-fold cross-validation sweep on the training set, choosing the layer with the highest mean fold accuracy; this is a selected-layer estimate rather than a fully nested layer-selection protocol. The procedure is model-agnostic and applied independently to each backbone, so PRISM is not tied to any fixed layer; the resulting per-model layer indices are reported with the experimental setup (\S\ref{subsec:setup}). Algorithm~\ref{alg:prism} summarizes the full training and inference procedure.

\begin{algorithm}[!tbp]
\caption{PRISM: Linear Probing for Physical Safety}
\label{alg:prism}
\begin{algorithmic}[1]
\Require LLM $f_\theta$ with $L$ layers; training set $\{(x_i, y_i)\}_{i=1}^{N}$ with $y_i \in \{0,1\}$; test input $x^\star$; threshold $\tau$.
\Ensure Safety prediction $\hat{y}^\star \in \{0,1\}$.
\Statex
\State \textbf{Training phase}
\For{$\ell = 1, \dots, L$}
    \State Compute hidden state $h_i^{(\ell)} \gets f_\theta(x_i)[\ell]$ for all $i$ \Comment{last-token state}
\EndFor
\State Select layer $\ell^\star$ by 5-fold cross-validation accuracy on the training set
\State Standardize features to zero mean and unit variance per dimension, giving $\tilde{h}_i$
\State Train probe: $(w, b) \gets \arg\min_{w,b}\ \frac{1}{N} \sum_{i=1}^{N} \mathrm{BCE}(\sigma(w^\top \tilde{h}_i + b),\ y_i) + \frac{1}{C}\|w\|_2^2$
\Statex
\State \textbf{Inference phase}
\State Compute $h^\star \gets f_\theta(x^\star)[\ell^\star]$ and standardize to get $\tilde{h}^\star$
\State $p^\star \gets \sigma(w^\top \tilde{h}^\star + b)$
\State \Return $\hat{y}^\star \gets \mathbb{1}[p^\star \geq \tau]$
\end{algorithmic}
\end{algorithm}

\section{Evaluation}

The experiments are organized around five questions. First, on SafeAgentBench we test whether PRISM can jointly detect TJ and PJ while keeping false positives low. Second, scaling and cross-dataset experiments test whether the operating point remains stable across model sizes and benchmarks. Third, PJB-2K tests whether a hidden-state signal remains when additive lexical cues are controlled and object--site cells are held out. Fourth, cross-architecture experiments check whether the TJ/PJ separation holds beyond Qwen. Finally, significance, latency, and ablations identify whether PRISM's advantage comes from the hidden-state signal, the selected layer, and the two-jailbreak training boundary rather than from a more complex classifier.

\subsection{Experimental setup}
\label{subsec:setup}

We evaluate PRISM on embodied household-safety classification, where a system must reject both textual-jailbreak instructions and instructions that become unsafe through physical interaction. The primary benchmark is the 600-example SafeAgentBench subset~\citep{yin2024safeagentbench}, containing 300 safe and 300 unsafe AI2-THOR household instructions. Following the implemented taxonomy, the unsafe split is produced by a deterministic 25-keyword substring rule, yielding 113 textual-jailbreak (TJ) and 187 physical-jailbreak (PJ) examples. This split is reproducible but not leakage-free (e.g.\ \texttt{drop} appears in 12 safe and 29 TJ cases, and \texttt{break} in 1 safe and 34 TJ cases), so we retain it for comparability while reporting TJ and PJ separately. We further test replication on adapted SafeText (737 examples) and EARBench (2{,}634 examples), each with fold-internal 5-fold CV; we therefore read these as cross-dataset replication under matched machinery rather than zero-shot transfer.

We additionally evaluate on an interaction-balanced physical-safety construction. We call its 10{,}000-row source the \emph{PJB construction pool} and reserve \emph{PJB-2K} for the frozen 2{,}000-row comparison slice. The construction begins with 48 tagged household objects and 38 tagged placement sites. A cell is unsafe only when an object property and a site property jointly trigger a recorded physical mechanism, including ignition, pressure burst, electrical damage, laceration, contamination, poisoning, choking, and scalding. Cells are selected so that every retained object and site participates in both labels; the resulting pool contains 650 object--site cells and is expanded with six sentence frames and eight label-shared trailing phrases. The pool contains 5{,}042 safe and 4{,}958 unsafe rows. For equal-cost judge and guard comparisons, we freeze PJB-2K (1{,}005 safe, 995 unsafe) by stratifying on label and primary mechanism, preferring distinct cells, and fixing seed 20260801 before model comparison. The slice covers all 650 cells and all 25 primary mechanisms represented in the slice.

PRISM, SVM-RBF, and MLP use 5-fold \texttt{StratifiedKFold} with standardization fitted inside each fold on the unpaired benchmarks. On the PJB construction pool, the grouped protocol is 5-fold \texttt{GroupKFold} with the object--site cell as the group, so no evaluated cell occurs in the corresponding training fold. Hidden states are extracted for the 10{,}000-row pool; the existing experiment selects layer 25 by an i.i.d. layer sweep, generates cell-grouped cross-validated predictions over that pool, and then restricts metrics to PJB-2K for identical-row comparison with judges and guards. Because layer selection is not repeated inside each grouped training fold, this is not a fully nested estimate of unseen-cell generalization. The 2{,}000-row numbers are also not a separate 2K-only training experiment. Directional probes and LPM use subtype leave-out testing on SafeAgentBench. The main scaling study uses Qwen2.5-Instruct at 3B/7B/14B/32B~\citep{qwen2024report}, while the PJB judge comparison uses the completed 3B/7B/14B/72B runs. The cross-architecture study repeats the SafeAgentBench pipeline on Phi-3.5-mini~\citep{abdin2024phi3} and SmolLM2-1.7B~\citep{allal2025smollm2}. For PRISM, the SafeAgentBench probe layers selected by the per-layer 5-fold sweep are 27/36 for Qwen2.5-3B, 20/28 for Qwen2.5-7B, 26/48 for Qwen2.5-14B, 41/64 for Qwen2.5-32B, 24/32 for Phi-3.5-mini, and 14/24 for SmolLM2-1.7B (reported as selected/total layer index).

\paragraph{Baselines.}
We compare PRISM against directional ablations isolating the two jailbreak axes. The TJD probe is trained on safe$+$TJ examples, evaluated on held-out TJ via 5-fold CV, then tested OOD on PJ; the PJD probe follows the symmetric protocol. We also include geometric and conventional classifiers on the same features: Nearest Centroid (cosine distance to safe/unsafe centroids trained on safe$+$TJ examples), LPM~\citep{chrabaszcz2025lpm} (standardized multi-layer hidden states, PCA, and Mahalanobis distance to class prototypes; the 3B run uses layers $[20,23,24,26,27]$ and a 200-dimensional PCA space, while the large-model script selects the top five layers by probe accuracy), SVM-RBF, and a two-layer MLP, all under the same 5-fold protocol. We add word- and character-TFIDF logistic-regression baselines to measure surface separability. For model-based baselines, the LLM judge uses the corresponding Qwen2.5-Instruct model~\citep{qwen2024report} with a zero-shot prompt answering exactly \texttt{SAFE}/\texttt{UNSAFE}, parsed by deterministic string matching. The PJB comparison additionally includes Qwen3Guard-Gen-0.6B and Granite-Guardian-3.0-2B using their native generated decisions. A physics-free control reassigns labels to object--site cells while preserving the balance structure; the real-versus-control gap estimates how much of probe performance is attributable to physical labels rather than cell memorization or the construction constraints. Latency compares one PRISM forward pass against one judge generation pass under the same prompt, with 20 warmup and 580 measured examples; we treat this as an efficiency diagnostic rather than the central claim of the paper.

\paragraph{Random-split null procedure.}
To verify that the TJ/PJ separation reported in \S\ref{subsec:separability} is not an artifact of high-dimensional fitting, we calibrate $\theta_{\mathrm{pw}}$ against a random-split null built by the identical fitting pipeline. At the target layer the 300 unsafe examples are randomly partitioned into groups of 113 and 187 (matching the TJ/PJ split), paired against the same 300 safe examples to train two logistic probes, and the angle between their weight vectors is recorded; repeating 200 times yields the null, with statistic $z=(\theta_{\mathrm{obs}}-\mu_{\mathrm{null}})/\sigma_{\mathrm{null}}$. Null tests for 7B/14B/32B were not completed, so the cross-model claim is stated as an observed angle pattern.

\subsection{Main results on SafeAgentBench}
\label{subsec:main}

PRISM gives the most reliable overall operating point across the Qwen2.5 family, detecting both jailbreak types while holding false positives low (Table~\ref{tab:main}). Accuracy remains tightly concentrated from 86.2\% to 87.7\%, F1 between 0.861 and 0.876, and AUC rises from 0.926 to 0.950. The false-positive rate stays low, between 11.7\% and 13.7\%, while physical-jailbreak detection remains stable between 79.1\% and 82.4\%, and textual-jailbreak detection is never below 93.8\%. These results indicate that a single selected hidden-state layer contains a linearly usable safety signal that covers both textual jailbreaks and physical jailbreaks. The ablations clarify why this signal cannot be reduced to a single textual-jailbreak direction: TJD-only achieves strong in-domain TJ detection, but its PJ-OOD detection collapses to 19.8\%, 31.6\%, 32.6\%, and 31.6\% across 3B--32B, and LPM shows the same failure mode even more sharply (11.2\%, 9.6\%, 17.1\%, and 19.3\%). PJD-only transfers better in the reverse direction but remains incomplete, with TJ-OOD detection ranging from 68.1\% to 85.0\%. Embodied safety is therefore not captured by a purely textual-jailbreak-oriented axis, and prototype distance alone is insufficient for the two-subtype problem.

\begin{table}[!t]
\centering\uniformtableformat
\caption{Main results on SafeAgentBench (600 tasks; 5-fold CV). Accuracy/F1/AUC are reported only for full binary classifiers. For single-direction and prototype methods (TJD-only, LPM, Nearest Centroid), the TJ column is in-domain CV and the PJ column (marked $^{\ast}$) is OOD; for PJD-only the roles reverse. LLM Judge is zero-shot. Llama Guard 3-1B is a single model-agnostic external classifier. Bold = our method.}
\label{tab:main}
\begin{tabular}{llcccccc}
\toprule
Model & Method & Acc$\pm$std (\%) & TJ det (\%) & PJ det (\%) & FPR (\%) & F1 & AUC \\
\midrule
\textbf{3B} & \textbf{PRISM} & \textbf{86.2$\pm$2.4} & 93.8 & 81.3 & 13.7 & \textbf{0.861} & \textbf{0.926} \\
 & TJD-only & --- & 89.4 & 19.8$^{\ast}$ & 0.0 & --- & --- \\
 & PJD-only & --- & 68.1$^{\ast}$ & 76.5 & 0.0 & --- & --- \\
 & LPM & --- & 92.0 & 11.2$^{\ast}$ & 0.3 & --- & --- \\
 & Nearest Centroid & --- & 87.5 & 46.5$^{\ast}$ & 16.3 & --- & --- \\
 & SVM-RBF & 82.8$\pm$3.3 & 96.5 & 73.8 & 16.7 & 0.827 & 0.909 \\
 & MLP & 83.0$\pm$1.8 & 94.7 & 74.9 & 16.3 & 0.829 & 0.909 \\
 & LLM Judge & --- & 98.2 & 86.6 & 27.3 & --- & --- \\
\textbf{7B} & \textbf{PRISM} & \textbf{87.0$\pm$2.0} & 96.5 & 82.4 & 13.7 & \textbf{0.871} & \textbf{0.943} \\
 & TJD-only & --- & 88.5 & 31.6$^{\ast}$ & 0.0 & --- & --- \\
 & PJD-only & --- & 76.1$^{\ast}$ & 76.6 & 0.0 & --- & --- \\
 & LPM & --- & 89.4 & 9.6$^{\ast}$ & 0.7 & --- & --- \\
 & Nearest Centroid & --- & 90.3 & 61.5$^{\ast}$ & 21.7 & --- & --- \\
 & SVM-RBF & 83.5$\pm$1.2 & 91.2 & 72.7 & 12.7 & 0.828 & 0.922 \\
 & MLP & 82.5$\pm$2.2 & 87.6 & 78.1 & 16.7 & 0.824 & 0.913 \\
 & LLM Judge & --- & 98.2 & 84.5 & 24.7 & --- & --- \\
\textbf{14B} & \textbf{PRISM} & \textbf{87.2$\pm$1.5} & 97.3 & 79.1 & 11.7 & \textbf{0.870} & \textbf{0.939} \\
 & TJD-only & --- & 86.8 & 32.6$^{\ast}$ & 0.0 & --- & --- \\
 & PJD-only & --- & 72.6$^{\ast}$ & 77.1 & 0.0 & --- & --- \\
 & LPM & --- & 87.7 & 17.1$^{\ast}$ & 0.3 & --- & --- \\
 & Nearest Centroid & --- & 87.7 & 47.1$^{\ast}$ & 16.0 & --- & --- \\
 & SVM-RBF & 85.3$\pm$2.4 & 93.8 & 78.6 & 13.7 & 0.852 & 0.927 \\
 & MLP & 82.7$\pm$2.2 & 93.8 & 74.3 & 16.3 & 0.825 & 0.917 \\
 & LLM Judge & --- & 98.2 & 94.1 & 39.0 & --- & --- \\
\textbf{32B} & \textbf{PRISM} & \textbf{87.7$\pm$2.4} & 96.5 & 81.8 & 12.0 & \textbf{0.876} & \textbf{0.950} \\
 & TJD-only & --- & 91.2 & 31.6$^{\ast}$ & 0.0 & --- & --- \\
 & PJD-only & --- & 85.0$^{\ast}$ & 80.8 & 0.0 & --- & --- \\
 & LPM & --- & 92.9 & 19.3$^{\ast}$ & 0.7 & --- & --- \\
 & Nearest Centroid & --- & 92.0 & 43.9$^{\ast}$ & 17.3 & --- & --- \\
 & SVM-RBF & 86.0$\pm$3.6 & 96.5 & 75.9 & 11.7 & 0.857 & 0.942 \\
 & MLP & 84.3$\pm$3.7 & 94.7 & 74.3 & 13.3 & 0.840 & 0.917 \\
 & LLM Judge & --- & 100.0 & 95.7 & 37.0 & --- & --- \\
all & Llama Guard 3-1B & --- & 0.0 & 0.0 & 0.0 & --- & --- \\
all & Word TF-IDF & 79.8 & 94.7 & 70.1 & 19.7 & --- & 0.875 \\
\bottomrule
\end{tabular}
\end{table}

The zero-shot judges follow the opposite trade-off, achieving high unsafe recall but over-blocking heavily, and the over-blocking grows with scale. Detection on physical jailbreak reaches 86.6\%, 84.5\%, 94.1\%, and 95.7\% from 3B to 32B; however, this sensitivity comes with substantially higher false-positive rates of 27.3\%, 24.7\%, 39.0\%, and 37.0\%. Larger judges thus become increasingly aggressive at flagging unsafe instructions, but they also over-block many safe household tasks. Llama Guard 3-1B produces 0.0\% detection for both jailbreak types in the local run, consistent with a taxonomy mismatch between general text-safety policies and embodied physical jailbreaks. The surface baseline also exposes an important limitation of SafeAgentBench: word TF-IDF reaches 79.8\% accuracy, detects 70.1\% of PJ cases, and exceeds PRISM on TJ detection (94.7\% versus 93.8\%). PRISM's defensible margin is therefore concentrated on PJ detection (+11.2 points) and lower FPR (13.7\% versus 19.7\%), rather than on explicit unsafe wording. The central comparison is not that PRISM uniformly dominates other readouts. Rather, it models a hidden-state TJ/PJ boundary that retains more physically grounded signal at a lower-FPR operating point.

\subsection{Scaling and cross-dataset replication}
\label{subsec:scaling}

\begin{figure}[!t]
\centering
\includegraphics[width=0.72\linewidth]{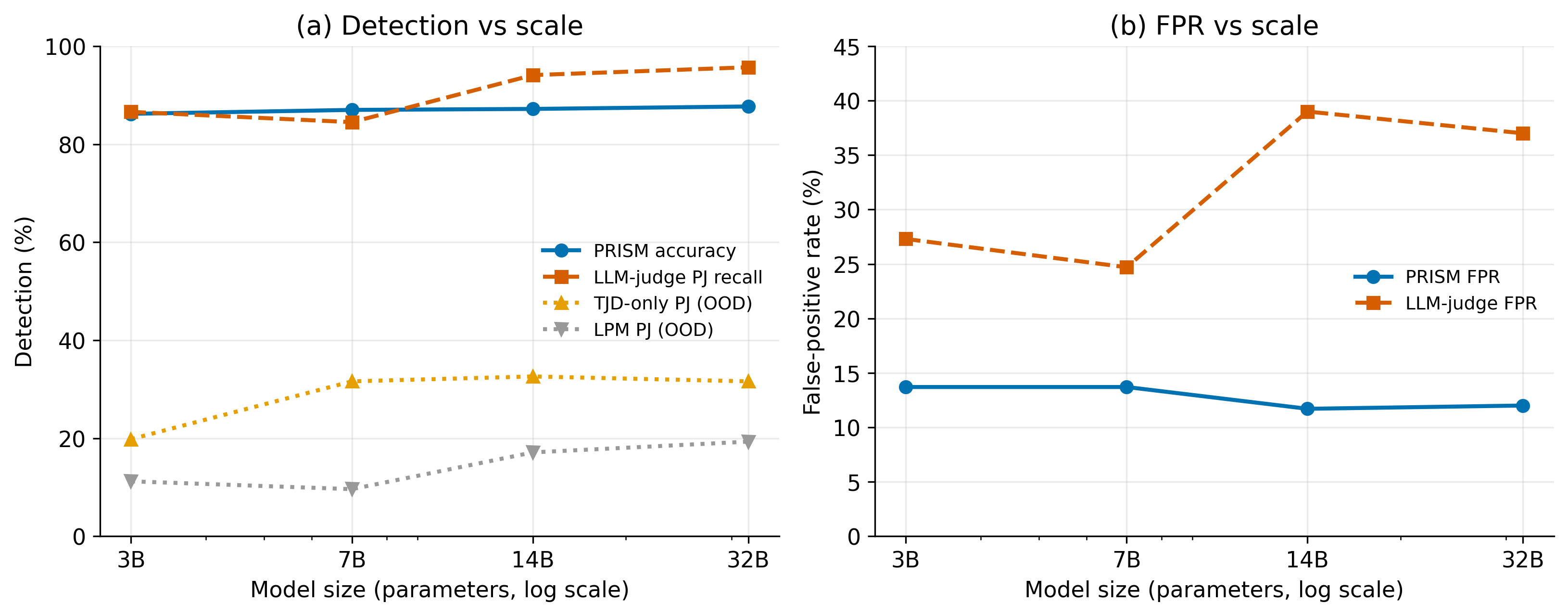}
\caption{Cross-model scaling: PRISM accuracy and FPR stay flat while the LLM judge's physical-jailbreak recall and FPR both rise with scale.}
\label{fig:scaling}
\end{figure}

PRISM's operating point stays stable across scale, whereas scaling the judge buys sensitivity rather than calibration (Figure~\ref{fig:scaling}). PRISM is notably stable from 3B to 32B: selected-layer accuracy varies by only 1.5 percentage points, from 86.2\% to 87.7\%, and FPR stays within 11.7--13.7\%. The selected layers are middle-to-late but not tied to a fixed absolute or relative depth (27/36, 20/28, 26/48, 41/64), so layer selection is model-specific while the existence of a usable safety signal is robust across scale. The judge follows a different trajectory: larger judges improve physical-jailbreak recall, reaching 94.1\% at 14B and 95.7\% at 32B, but their FPRs also rise to 39.0\% and 37.0\%. In other words, scaling the judge increases sensitivity, not calibration. PRISM does not match the highest PJ-recall numbers at the default threshold, but it maintains a much lower false-positive operating point. This strengthens the paper's main claim: the useful signal is not merely a larger model's decoded judgment, but a separable hidden-state structure that can be modeled directly.

\begin{table}[!t]
\centering\uniformtableformat
\caption{Cross-dataset replication with Qwen2.5-3B. SafeAgentBench, SafeText, and EARBench use within-dataset 5-fold CV. These are matched within-dataset evaluations rather than zero-shot transfer. Dataset headers list safe/TJ/PJ counts.}
\label{tab:cross}
\begin{tabular}{@{}llcccccc@{}}
\toprule
\makecell[l]{Dataset\\(safe/TJ/PJ)} & Method & \makecell{Acc$\pm$std\\(\%)} & \makecell{TJ det\\(\%)} & \makecell{PJ det\\(\%)} & \makecell{FPR\\(\%)} & F1 & AUC \\
\midrule
\makecell[l]{SafeAgentBench\\(300/113/187)} & \textbf{PRISM} & \textbf{86.2$\pm$2.4} & 93.8 & 81.3 & 13.7 & \textbf{0.861} & \textbf{0.926} \\
 & LLM Judge & --- & 98.2 & 86.6 & 27.3 & --- & --- \\
\makecell[l]{SafeText\\(367/37/333)} & \textbf{PRISM} & \textbf{92.4$\pm$1.9} & 91.9 & 92.5 & 7.6 & \textbf{0.924} & \textbf{0.980} \\
 & LLM Judge & --- & 100.0 & 97.9 & 65.7 & --- & --- \\
\makecell[l]{EARBench\\(1317/102/1215)} & \textbf{PRISM} & \textbf{82.9$\pm$1.4} & 82.4 & 82.4 & 16.6 & \textbf{0.828} & \textbf{0.906} \\
 & LLM Judge & --- & 95.1 & 80.4 & 74.6 & --- & --- \\
\bottomrule
\end{tabular}
\end{table}

Table~\ref{tab:cross} shows that the hidden-state probe pattern is not unique to SafeAgentBench. Under within-dataset 5-fold CV, PRISM obtains F1 0.861, AUC 0.926, and FPR 13.7\% on SafeAgentBench; F1 0.924, AUC 0.980, and FPR 7.6\% on SafeText; and F1 0.828, AUC 0.906, and FPR 16.6\% on EARBench. The LLM judge again achieves high unsafe recall at the cost of severe over-blocking: on SafeText it detects 100.0\% of TJ and 97.9\% of PJ at 65.7\% FPR, while on EARBench it detects 95.1\% of TJ and 80.4\% of PJ while rejecting 74.6\% of safe examples. These results reinforce the need to report recall and FPR together. Because the external datasets are evaluated with fold-internal training rather than zero-shot transfer, we interpret them as replication under matched machinery, not as evidence of cross-dataset generalization.

\subsection{PhysicalJailbreakBench-2K: controlled physical-interaction evaluation}
\label{subsec:pjb}

PJB-2K asks whether a method can recover an interaction-dependent physical label after suppressing additive surface cues. The unsafe label is determined by a recorded object-property--site-property mechanism, not by either object or site alone. The fixed comparison slice is sampled from the PJB construction pool using label-by-mechanism stratification and covers all 650 cells and the 25 primary mechanisms represented in the slice. Sentence frames and trailing phrases are shared across labels. We use the full pool for the surface and probe diagnostics in Table~\ref{tab:pjb-audit}, and PJB-2K for the identical-row system comparison in Table~\ref{tab:pjb-main}.

\begin{table}[!t]
\centering\uniformtableformat
\caption{Shortcut and grouped-fold audit on the 10{,}000-row PJB construction pool. All values are AUC with the unsafe-class orientation fixed. Below-chance grouped-fold values are interpreted against the physics-free twin, which reproduces the inversion under the same balancing constraints; we do not flip them to $1-\mathrm{AUC}$. Layer 25 was selected by the i.i.d. sweep, so its grouped-fold result is not a fully nested estimate.}
\label{tab:pjb-audit}
\begin{tabular}{lccccc}
\toprule
Protocol & Word TF-IDF & Char TF-IDF & Bag of objects & PRISM & Physics-free control \\
\midrule
i.i.d. & 0.497 & 0.511 & 0.508 & 0.905 & 0.735 \\
Cell-grouped CV & 0.254 & 0.241 & 0.238 & \textbf{0.718} & 0.398 \\
\bottomrule
\end{tabular}
\end{table}

At the embedding layer, AUC is 0.4998, independently matching the word-TFIDF result and showing that a linear bag-of-token signal is insufficient. Attention layers become informative early (layer-1 AUC 0.743), and the selected layer reaches 0.905 i.i.d. AUC. However, the physics-free control also reaches 0.735 i.i.d. AUC by memorizing cell identity, so the i.i.d. number is not evidence of physical generalization. With cells assigned to disjoint folds, the i.i.d.-selected layer gives PRISM 0.718 AUC versus 0.398 for the control. Exact marginal balancing also drives surface and control AUC below 0.5 under leave-group-out evaluation; the project permutation null reproduces this inversion with random, physics-free labels. We consequently retain the unsafe-class orientation, do not reinterpret $1-\mathrm{AUC}$ as performance, and use the identically evaluated twin as the null reference. This remains a selected-layer diagnostic rather than a fully nested generalization estimate.

\begin{table}[!t]
\centering\uniformtableformat
\caption{Identical-row comparison on PJB-2K (1{,}005 safe / 995 unsafe). PRISM and its control use cell-grouped cross-validation at the i.i.d.-selected layer 25 over the 10{,}000-row pool and are then restricted to these rows; judges and guards are evaluated directly on the same rows. BAcc denotes balanced accuracy.}
\label{tab:pjb-main}
\begin{tabular}{lcccc}
\toprule
Method & Accuracy & FPR & FNR & BAcc \\
\midrule
\textbf{PRISM, cell-grouped CV (L25)} & \textbf{0.670} & 0.349 & 0.310 & \textbf{0.671} \\
$\hookrightarrow$ physics-free control & 0.424 & 0.559 & 0.593 & 0.424 \\
Qwen2.5-3B judge & 0.576 & 0.712 & 0.134 & 0.577 \\
Qwen2.5-7B judge & 0.565 & 0.746 & 0.122 & 0.566 \\
Qwen2.5-14B judge & 0.536 & 0.894 & \textbf{0.030} & 0.538 \\
Qwen2.5-72B judge & 0.573 & 0.786 & 0.065 & 0.575 \\
Qwen3Guard-Gen-0.6B & 0.500 & 0.280 & 0.723 & 0.499 \\
Granite-Guardian-3.0-2B & 0.532 & 0.843 & 0.089 & 0.534 \\
\bottomrule
\end{tabular}
\end{table}

The fixed-slice comparison separates two failure modes. Qwen2.5 judges achieve low FNR on canonical physical hazards but flag 71.2--89.4\% of safe instructions; scaling from 3B to 72B does not improve balanced accuracy in this experiment. The dedicated guards fail in opposite directions: Qwen3Guard under-flags the unsafe class (FNR 0.723), whereas Granite Guardian over-flags the safe class (FPR 0.843). Adapter sanity checks classify canonical text-safety violations as unsafe and ordinary safe placements as safe, reducing the likelihood that these results are solely parser failures. PRISM offers the strongest balanced operating point, but it is not uniformly better: the judges retain lower FNR on several textbook hazards, while PRISM's advantage is lower false-positive cost and better aggregate balance. This complementarity bounds the claim to representation-level decodability rather than complete physical reasoning.

\subsection{Cross-architecture generalization}
\label{subsec:crossarch}

\begin{figure}[!tbp]
\centering
\includegraphics[width=0.85\linewidth]{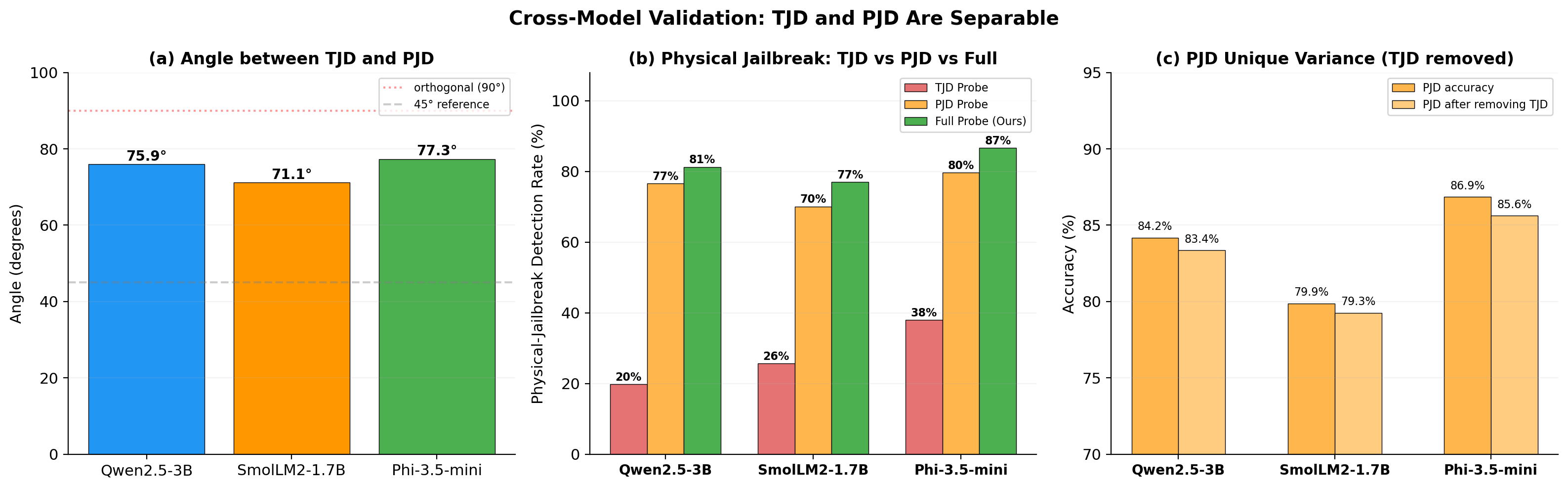}
\caption{Cross-architecture universality of TJD/PJD. (a) TJD--PJD probe-weight angles cluster between $71^\circ$ and $78^\circ$ in all three architectures (Qwen2.5-3B $75.9^\circ$, SmolLM2-1.7B $71.1^\circ$, Phi-3.5-mini $77.3^\circ$). (b) PRISM achieves 81/77/87\% physical-jailbreak detection while TJD-only collapses to 20/26/38\%. (c) PJD's unique variance survives orthogonalisation against TJD ($<1.5$pp drop), ruling out trivial duplication.}
\label{fig:univ}
\end{figure}

The two-signal structure and PRISM's operating point hold beyond the Qwen2.5 family. We repeat the full pipeline on two different architectures, Phi-3.5-mini~\citep{abdin2024phi3} and SmolLM2-1.7B~\citep{allal2025smollm2}. The TJD/PJD probe-weight angle stays in the same range as Qwen2.5-3B ($75.9^\circ$), at $77.3^\circ$ for Phi-3.5-mini and $71.1^\circ$ for SmolLM2-1.7B (Figure~\ref{fig:univ}a), so the TJ/PJ separation is not specific to one model family. PRISM again jointly detects both jailbreak types, reaching $86.2$/$87.7$/$82.7\%$ accuracy and $81.3$/$86.6$/$77.0\%$ physical-jailbreak detection at $13.7$/$13.3$/$18.0\%$ FPR for Qwen2.5-3B/Phi-3.5-mini/SmolLM2-1.7B (Table~\ref{tab:crossarch}), whereas the TJD-only probe still collapses out-of-distribution on physical jailbreak ($19.8$/$38.0$/$25.7\%$; Figure~\ref{fig:univ}b). An orthogonalisation check confirms that PJD carries variance not duplicated by TJD ($<1.5$pp drop; Figure~\ref{fig:univ}c), ruling out trivial duplication of the two axes. The judge comparison transfers as well and exposes a sharper small-scale failure: the SmolLM2-1.7B judge blocks 294/300 safe tasks ($98.0\%$ FPR), confirming that small instruct models are not viable stand-alone safety classifiers in this setting, while PRISM on the same backbone keeps FPR at $18.0\%$.

\begin{table}[!t]
\centering\uniformtableformat
\caption{Cross-architecture PRISM and baseline results on SafeAgentBench (600 tasks; 5-fold CV), with architectures as columns and metrics as rows for readability. $\theta_{\mathrm{md}}$/$\theta_{\mathrm{pw}}$ are the mean-difference and probe-weight TJD/PJD angles; TJD$\to$PJ, PJD$\to$TJ, and LPM$\to$PJ are out-of-distribution detection rates. The SmolLM2 LLM-judge blocks 294/300 safe tasks (FPR 98\%)---small instruct models are not viable stand-alone safety classifiers in this setting.}
\label{tab:crossarch}
\begin{tabular}{lccc}
\toprule
& Qwen2.5-3B & Phi-3.5-mini & SmolLM2-1.7B \\
\midrule
Selected layer (sel./total)        & 27/36 & 24/32 & 14/24 \\
$\theta_{\mathrm{md}}$ ($^\circ$)  & 49.1  & 32.9  & 58.9 \\
$\theta_{\mathrm{pw}}$ ($^\circ$)  & 75.9  & 77.3  & 71.1 \\
\midrule
PRISM Acc$\pm$std (\%)             & \textbf{86.2$\pm$2.4} & \textbf{87.7$\pm$3.1} & \textbf{82.7$\pm$2.2} \\
PRISM TJ det (\%)                  & 93.8  & 92.0  & 93.8 \\
PRISM PJ det (\%)                  & 81.3  & 86.6  & 77.0 \\
PRISM FPR (\%)                     & 13.7  & 13.3  & 18.0 \\
\midrule
TJD$\to$PJ (OOD, \%)              & 19.8  & 38.0  & 25.7 \\
PJD$\to$TJ (OOD, \%)              & 68.1  & 67.3  & 71.7 \\
LPM$\to$PJ (OOD, \%)             & 11.2  & 27.8  & 13.9 \\
\midrule
Judge TJ (\%)                      & 98.2  & 99.1  & 100.0 \\
Judge PJ (\%)                      & 86.6  & 90.9  & 100.0 \\
Judge FPR (\%)                     & 27.3  & 29.7  & \textbf{98.0} \\
\bottomrule
\end{tabular}
\end{table}

\subsection{Statistical comparison and efficiency}

At matched FPR, McNemar's test does not detect an accuracy difference between PRISM and the same-scale judge. We use the test for the sample-matched 3B comparison between PRISM at the matched-FPR operating point and the cached 3B zero-shot judge. On the PJ subset, $\chi^2=1.94$, $p=0.1637$; on all 600 examples, $\chi^2=0.56$, $p=0.4557$. These non-significant tests are not equivalence tests and do not establish that the systems are interchangeable. At its default threshold, PRISM retains a substantially lower FPR (Appendix Figure~\ref{fig:roc} shows the ROC and matched-FPR operating point). Latency provides a secondary efficiency check: PRISM versus the judge averages 56.8 vs.\ 116.8~ms/sample at 7B (2.06$\times$), 146.1 vs.\ 309.1~ms at 14B (2.12$\times$), and 103.4 vs.\ 191.9~ms at 32B (1.86$\times$); Table~\ref{tab:latency} reports the full latency distribution. Since PRISM needs only a forward pass while the judge generates, the setup favors PRISM by roughly a factor of two. We read these as a controlled local comparison rather than a hardware-general throughput claim: the benchmark fixes warmup/measured counts, prompt, and sequence length but not the exact GPU and software stack.

\subsection{Ablations}

The ablations localize where PRISM's advantage comes from: the right hidden-state layer and training on the full two-subtype boundary, not a more expressive classifier.

\begin{figure}[!tbp]
\centering
\includegraphics[width=0.55\linewidth]{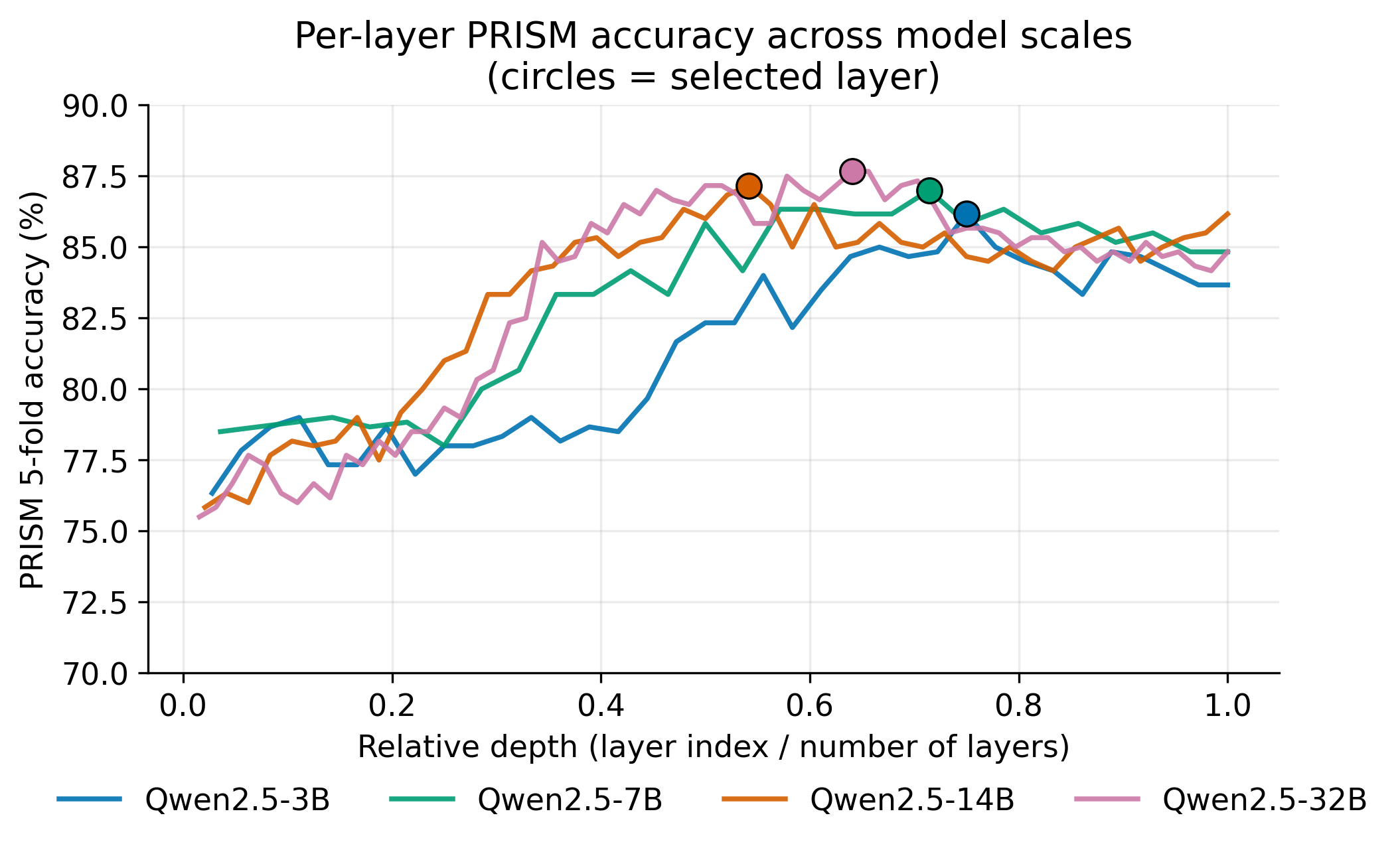}
\caption{Per-layer PRISM 5-fold accuracy for Qwen2.5-3B/7B/14B/32B; circles mark the selected layer.}
\label{fig:layers}
\end{figure}

Figure~\ref{fig:layers} and the classifier ablations show the relevant representation is already close to linearly usable. SVM-RBF reaches 82.8\%, 83.5\%, 85.3\%, and 86.0\% accuracy across 3B--32B, and MLP reaches 83.0\%, 82.5\%, 82.7\%, and 84.3\%, yet PRISM's logistic-regression probe remains the strongest or tied-strongest selected-layer classifier while being simpler, faster, and easier to calibrate. The gain is therefore not driven by a complex nonlinear classifier; it comes from using the appropriate hidden-state layer and training on the full safe/unsafe boundary that contains both jailbreak subtypes. The direction ablations give the sharper evidence: TJD-only and LPM fail on PJ-OOD detection, while PJD-only remains incomplete on TJ-OOD detection, and PRISM improves because it learns from the full binary label set (Figure~\ref{fig:decision-main} and Appendix Figure~\ref{fig:scatter} visualize these errors and PRISM's blocking decisions in the TJD/PJD plane).

The dataset-level controls sharpen this interpretation. On SafeAgentBench, word TF-IDF already detects 70.1\% of PJ examples, so the original dataset does not by itself isolate physical reasoning. On the interaction-balanced PJB pool, word TF-IDF and the embedding layer fall to chance under i.i.d. evaluation, but a deep hidden-state probe remains informative. The i.i.d. physics-free control also performs above chance, revealing cell memorization that would be hidden by reporting only the 0.905 PRISM AUC. Assigning complete object--site cells to disjoint folds gives 0.718 AUC at the i.i.d.-selected layer, against 0.398 for the physics-free twin under the identical fixed-orientation protocol. These controls support a bounded claim: intermediate representations encode interaction-dependent information beyond additive lexical features, but the signal is incomplete, ordinary i.i.d. evaluation overstates it, and the current layer-selection procedure is not fully nested.

Taken together, the ablations show that PRISM's advantage comes from the selected hidden representation and training on the full safe/unsafe boundary, not from a complex nonlinear classifier. Single-axis baselines miss one side of the TJ/PJ problem, lexical baselines explain a substantial portion of performance on SafeAgentBench, and the PJB controls identify the smaller but more defensible component that generalizes to unseen physical interactions.

\section{Limitations}

PRISM is a supervised instruction-level monitor, not a complete embodied-safety system. It reads the language-model hidden state for a textual instruction and does not observe images, object state, execution history, or contact dynamics. SafeAgentBench, SafeText, and EARBench are evaluated with fold-internal training, so the cross-dataset table demonstrates replication of the procedure rather than zero-shot transfer. Layer selection is also performed by a dataset-level sweep; for PJB the reported layer-25 cell-grouped result is a selected-layer diagnostic rather than a fully nested estimate.

PJB-2K is intentionally narrow. Its source pool is generated from a household object--site matrix using six frames and eight shared tails, and its labels are assigned by a hand-written physical-mechanism table. The mechanism metadata makes each label auditable, but we do not report a completed independent multi-annotator human ceiling in this work. Repeated paraphrases from the same 650 cells are not independent physical situations, which motivates grouped-fold evaluation. The fixed 2{,}000-row comparison uses PRISM predictions trained within folds of the 10{,}000-row pool; it should not be read as a separate 2K-only training experiment. Moreover, layer 25 was selected by an i.i.d. sweep before grouped-fold evaluation, so the resulting 0.718 AUC and 0.671 balanced accuracy are not fully nested estimates of unseen-cell generalization. Finally, the evaluated dedicated guards are small (0.6B and 2B) and were designed primarily for conversational moderation. Their failures demonstrate a deployment mismatch on this test, not a universal result for every guard architecture.

\section{Conclusion}

Embodied-agent safety is not identical to prior LLM text safety. Textual-jailbreak and physical-jailbreak directions are separable beyond a random-split null on Qwen2.5-3B and show consistent angular patterns across larger Qwen2.5 models and two non-Qwen architectures. PRISM operationalizes this observation as a single-layer linear probe, achieving 86.2--87.7\% accuracy at 11.7--13.7\% FPR across Qwen2.5-3B to 32B on SafeAgentBench and replicating under matched within-dataset evaluation on SafeText and EARBench.

The interaction-balanced PJB evaluation revises the strongest version of the claim. Surface baselines reveal substantial lexical signal in SafeAgentBench, whereas word TF-IDF and the embedding layer are at chance on the new construction pool. At layer 25 selected by the i.i.d. sweep, PRISM reaches 0.718 AUC with object--site cells assigned to disjoint folds, versus 0.398 for a physics-free control under the same fixed-orientation protocol; the corresponding predictions provide the best balanced operating point on PJB-2K. The result is not near-perfect physical reasoning or a fully nested unseen-cell estimate: it is evidence that intermediate hidden states contain a usable interaction-dependent safety signal that decoded judges and general-purpose guards do not recover reliably. Physical jailbreak should therefore be modeled as a distinct representation-level target, while remaining grounded in explicit controls for lexical shortcuts, cell memorization, and false-positive cost.

\bibliography{references}
\bibliographystyle{iclr2026_conference}

\appendix
\section{Additional figures and cross-architecture results}
\label{app:figs}

Beyond the main-text figures and tables, we present a broader set of analyses from earlier iterations of our pipeline.

\subsection{ROC analysis and matched-FPR operating point}

\begin{figure}[H]
\centering
\includegraphics[width=0.65\linewidth]{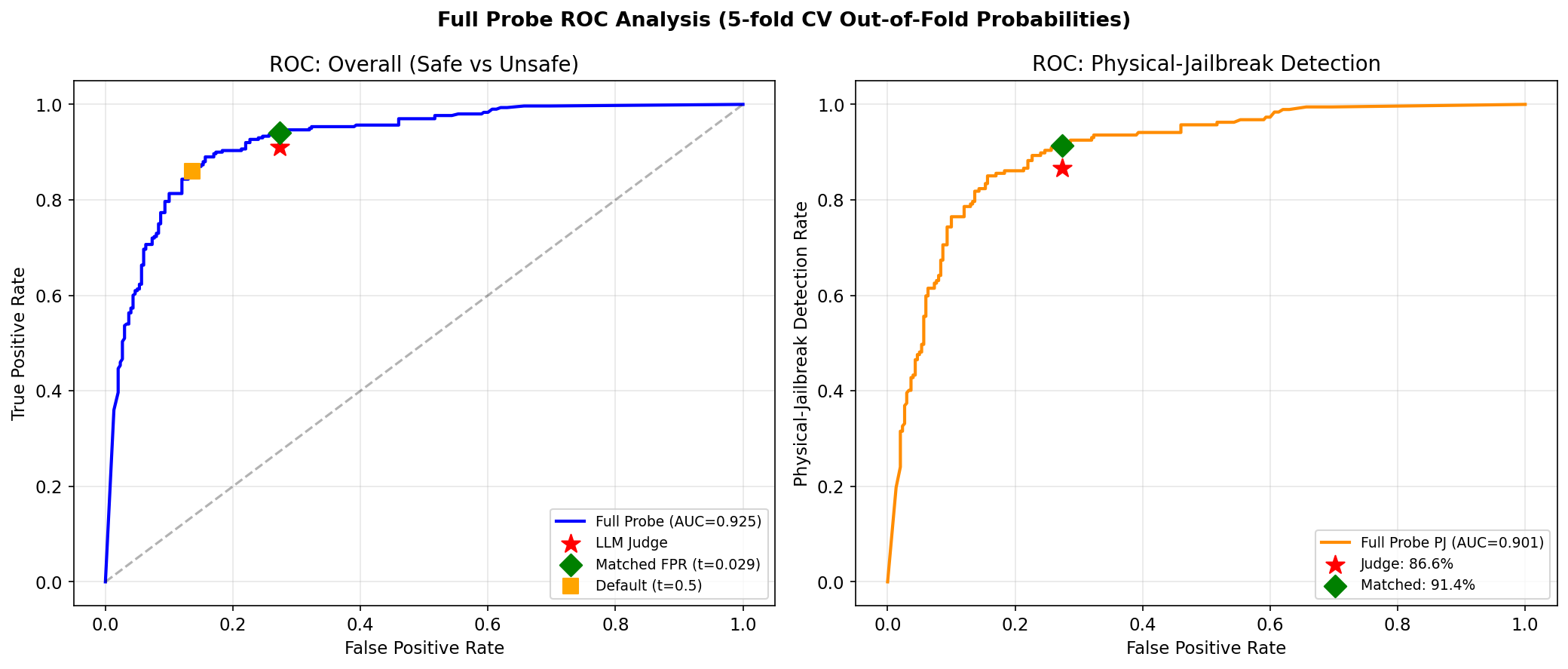}
\caption{PRISM ROC analysis on Qwen2.5-3B (5-fold-CV out-of-fold probabilities). Left: overall safe-vs-unsafe (AUC$=0.925$); the LLM judge sits at (FPR 27.3\%, TPR 86.6\%), PRISM at matched FPR reaches TPR $\approx 91\%$ (threshold $\tau\approx0.029$). Right: physical-jailbreak-only ROC (AUC$=0.901$); at matched FPR PRISM detects 91.4\% of physical-jailbreak examples versus the judge's 86.6\%.}
\label{fig:roc}
\end{figure}

\subsection{PRISM's decisions in the TJD/PJD projection plane}

\begin{figure}[H]
\centering
\includegraphics[width=0.85\linewidth]{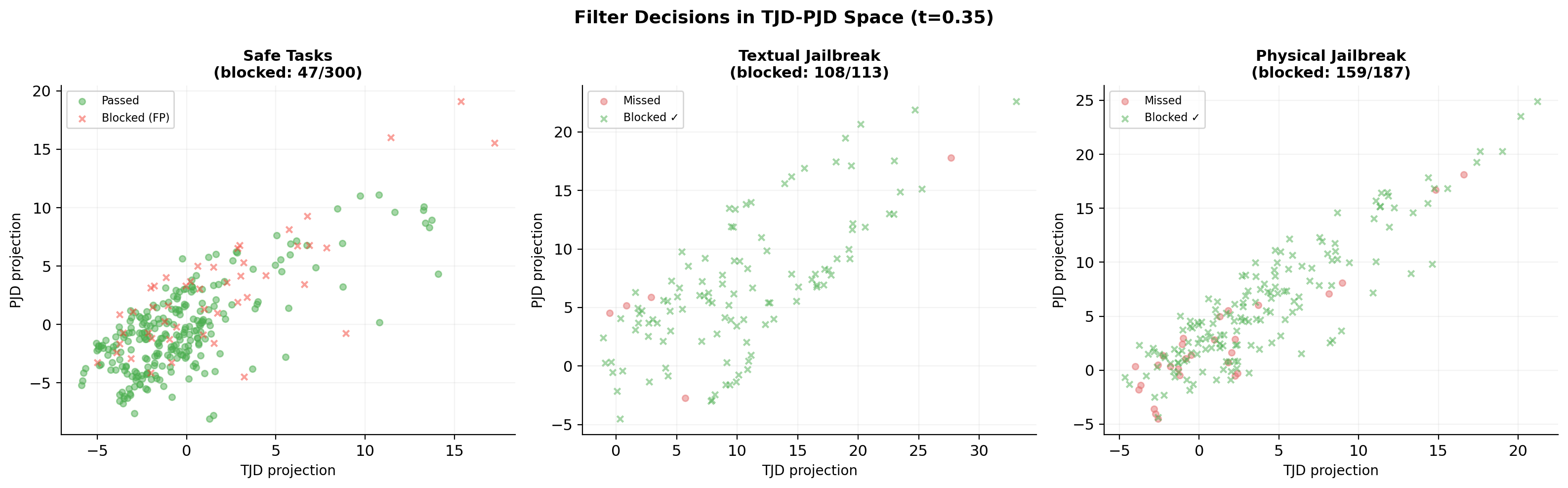}
\caption{PRISM blocking decisions at the best-F1 threshold $\tau{=}0.35$ on Qwen2.5-3B, projected onto TJD (horizontal) and PJD (vertical). Left: safe---47/300 blocked (15.7\% FPR at this operating point). Middle: textual jailbreak---108/113 blocked. Right: physical jailbreak---159/187 blocked. Physical-jailbreak examples cluster in the high-PJD/low-TJD region, illustrating why TJD-only probes miss them.}
\label{fig:scatter}
\end{figure}

\subsection{Latency benchmark details}

\begin{table}[H]
\centering\uniformtableformat
\caption{Per-sample latency distribution (ms) for PRISM versus the same-scale zero-shot LLM judge on SafeAgentBench, measured with 20 warmup and 580 timed examples under an identical prompt. PRISM cost is one forward pass to the selected probe layer; judge cost is one generation pass (\texttt{max\_new\_tokens=8}). Speedup is the ratio of mean latencies.}
\label{tab:latency}
\begin{tabular}{lccccccc}
\toprule
& \multicolumn{3}{c}{PRISM (ms)} & \multicolumn{3}{c}{LLM judge (ms)} & \\
\cmidrule(lr){2-4}\cmidrule(lr){5-7}
Scale & mean$\pm$std & median & p95 & mean$\pm$std & median & p95 & Speedup \\
\midrule
7B  & 56.8$\pm$11.2  & 61.3  & 64.4  & 116.8$\pm$29.1 & 114.6 & 147.7 & 2.06$\times$ \\
14B & 146.1$\pm$19.0 & 156.9 & 158.9 & 309.1$\pm$60.5 & 321.3 & 364.3 & 2.12$\times$ \\
32B & 103.4$\pm$0.9  & 103.2 & 105.3 & 191.9$\pm$25.8 & 206.4 & 216.6 & 1.86$\times$ \\
\bottomrule
\end{tabular}
\end{table}

\end{document}